\documentclass[conference]{IEEEtran}
\IEEEoverridecommandlockouts
\usepackage{graphicx} 
\usepackage[dvipsnames]{xcolor}
\usepackage{algorithm}
\usepackage{cite}
\usepackage{algpseudocode}
\usepackage{booktabs} 
\usepackage{subcaption}
\usepackage{amsmath}
\usepackage[font=small,belowskip=-1.5pt]{caption} 
\makeatletter
\let\NAT@parse\undefined
\makeatother
\usepackage{hyperref}
\hypersetup{
    colorlinks=true,
    linkcolor=blue,
    filecolor=magenta,      
    urlcolor=cyan,
    pdftitle={DREAM},
    pdfpagemode=FullScreen,
    }

\graphicspath{{images/}}

\title{DREAM: Decentralized Reinforcement Learning for Exploration and Efficient Energy Management in Multi-Robot Systems}
\author{Dipam Patel, Phu Pham, Kshitij Tiwari and Aniket Bera%
\thanks{Department of Computer Science, Purdue University, USA
    {\tt\small \{dipam,pham84,tiwarik,ab\}@purdue.edu}}\\
\thanks{Supplemental Version: {\small \href{https://ideas.cs.purdue.edu/research/dream}{ideas.cs.purdue.edu/research/dream}}}
\vspace{-10pt}
}

\begin{document}

\maketitle

\begin{abstract}
\label{abstract}
Resource-constrained robots often suffer from energy inefficiencies, underutilized computational abilities due to inadequate task allocation, and a lack of robustness in dynamic environments, all of which strongly affect their performance. This paper introduces \textit{DREAM} - Decentralized Reinforcement Learning for Exploration and Efficient Energy Management in Multi-Robot Systems, a comprehensive framework that optimizes the allocation of resources for efficient exploration. It advances beyond conventional heuristic-based task planning as observed conventionally. The framework incorporates Operational Range Estimation using Reinforcement Learning to perform exploration and obstacle avoidance in unfamiliar terrains. DREAM further introduces an Energy Consumption Model for goal allocation, thereby ensuring mission completion under constrained resources using a Graph Neural Network. This approach also ensures that the entire Multi-Robot System can survive for an extended period of time for further missions compared to the conventional approach of randomly allocating goals, which compromises one or more agents. Our approach adapts to prioritizing agents in real-time, showcasing remarkable resilience against dynamic environments. This robust solution was evaluated in various simulated environments, demonstrating adaptability and applicability across diverse scenarios. We observed a substantial improvement of about 25\% over the baseline method, leading the way for future research in resource-constrained robotics.

\end{abstract}

\section{Introduction}\label{intro}
In the domain of robotics, addressing the unique challenges and opportunities presented by resource-constrained robots is of paramount significance. Such robots are expected to employ their computational ability and sensor capabilities to execute tasks in the most energy-efficient manner. Integral to these challenges are issues related to energy consumption, path planning, and coordinated decision-making, especially in multi-robot scenarios. As robotic systems are increasingly applied in diverse fields, ranging from environmental monitoring, natural disasters \cite{7986575}, and search \& rescue missions \cite{article}, solving these resource constraints has become crucial.


Constructing an energy consumption model that comprehensively factors in the variables impacting energy use within a robot poses significant challenges. This complexity arises from the necessity to conduct precise system identification procedures and accurately quantify each energy expense. The operational range estimation of a robot refers to the distance within which it can function given its resource constraints. Energy should be perceived as a predetermined resource constraint that optimizes the mission execution within these defined limitations. This model, therefore, needs to be adaptable, and capable of responding to real-time changes.

\begin{figure}[H]
  \centering
  \includegraphics[width=1\linewidth]{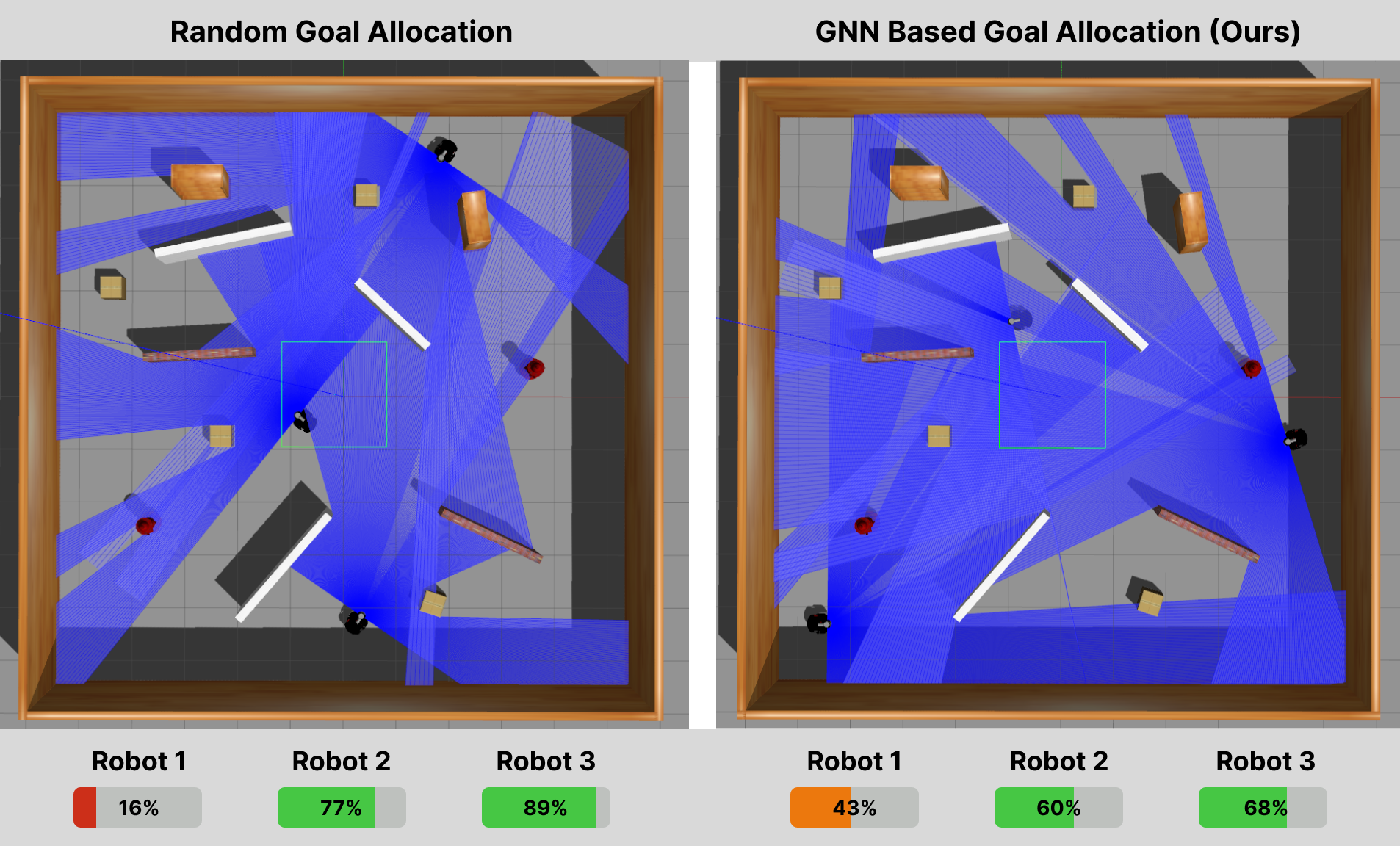}
  \caption{\textit{Three Robots (in black) navigating to their respective Goal positions using the Refined TD3 (RTD3) Model. Upon initiation, the GNN model uses all the Robots' states to do goal allocation, which accounts for minimizing the cumulative energy used by the system at the end of the mission.}}
  \label{fig:}
\end{figure}

In the context of co-operative Multi-Robot Systems (MRS), the necessity for efficient resource management becomes more complex and crucial. Compared to a single robot, an MRS setup offers improved reliability and scalability. It can cover large areas and complete tasks quickly and efficiently, making it suitable for applications in diverse fields. However, this also leads to challenges associated with goal allocation and trajectory coordination under constrained resources. 

Efficient task allocation in an MRS involves assigning suitable roles to individual robots based on their specific capabilities and mission requirements. Trajectory coordination, on the other hand, involves planning the paths of individual robots in a way that they can collaborate effectively without interfering with each other's tasks. This requires careful planning and real-time adaptability to avoid collisions and ensure smooth cooperation among the robots with limited resources. This makes decentralized navigation essential, where each robot is equipped to make independent decisions based on local information. It can adjust its path dynamically, responding to real-time changes in the environment.

We propose a Refined Twin Delayed Deep Deterministic Policy Gradient (RTD3) based model for obstacle avoidance and exploration for Multi-Robot Systems. We also propose a Graph Neural Network (GNN) based model that leverages real-time operational data for instantaneous goal allocation. In summary, our model prioritizes developing robust path-planning mechanisms under strict energy constraints which drives the development of DREAM. The key contributions of our work can be summarized as follows:
\begin{itemize}
    \item Introduced a \textbf{Refined TD3} structure, leveraging the \textbf{Reward Categorized Replay Buffer}, resulting in a \textbf{75\%} reduction in model parameters.
    \item Developed a GNN-based \textbf{Energy Management Model} for adaptive mission planning based on real-time energy availability, enhancing mission success and system lifespan.
    \item Expanded navigation capabilities for multi-agent, goal-driven exploration and collaborative mapping, using a single agent-goal pair for training.
    \item Amplified the model's versatility across varying environments, regardless of the robot-goal pair count, leveraging the benefits of \textbf{Curriculum Learning}.
\end{itemize}


\section{Related Works}
\label{related}

The increasing complexity and scalability of problems for robotics have necessitated the deployment of MRS, with agents discovering solutions using learning mechanisms \cite{MultiRL,8352646,s22218099}. Such systems have gained tremendous attention in recent years, with applications ranging from complex system modeling to smart grids and computer networks. Nonetheless, MRS presents inherent challenges, including agent coordination, security, and task allocation \cite{8352646,s22218099}.

Reinforcement learning techniques have been extensively used in multiagent systems. For example, \cite{MultiRL} provides a comprehensive survey of multi-agent reinforcement learning. Likewise, \cite{LoweWTHAM17} and \cite{Yang} explore deep reinforcement learning methods for multi-agent domains, with the latter introducing the Mean Field Reinforcement Learning approach to address the scalability issue. Interestingly, \cite{Yang} also establishes a relationship between the learning of an individual agent's optimal policy and the dynamics of the population, highlighting the mutual reinforcement between the two.

In the context of swarm robotics, inspired by natural swarms' collaborative intelligence, \cite{RAKESH2022108020} investigates a fault-tolerant pattern formation algorithm. The paper sheds light on how different agents can form a geometric pattern and maintain it, even in the presence of faulty agents. In a related study, \cite{8511420} proposes a technique for the identification of biased-measurement agents in a network of mobile robots, highlighting the impact of errors on system performance.

Energy efficiency and optimization have also been a significant focus in multi-agent systems and robotics. For instance, \cite{7101619} offers an energy-aware path planning algorithm for UAVs, while \cite{energies} examines energy optimization for optimal motion planning for a dual-arm industrial robot. Further, \cite{su141610056} describes an energy management system for mobile robots used for search and rescue that integrates a battery and supercapacitor to manage power sharing.

Effective inter-agent communication is pivotal in decentralized multi-robot coordination. Traditional methods have employed discrete communication strategies, such as signal binarization \cite{foerster2016learning}. The infusion of attention mechanisms into GNNs offers a promising direction to process information-dense graphs \cite{veličković2018graph}. While attention on static graphs has shown potential, its efficacy in dynamic multi-agent graphs remains to be fully explored \cite{liu2020when2com}.

Deep learning advancements have enabled robots to anticipate their environments and conduct directed explorations. Notable works include exploring navigation via deep Q-learning \cite{7784001}, utilizing RGB images to anticipate maps and actions \cite{chaplot2020learning}, \cite{ramakrishnan2020occupancy}, and using learned predictions to map surroundings \cite{8793769}.

Our research extends the work of Cimurs et al. \cite{9645287} for Multi-Robot Systems under strict resource constraints. Our proposal integrates a streamlined motion policy based on raw Lidar data with a holistic global navigation strategy. Our approach not only aims for goal-driven exploration in uncharted terrains along with obstacle avoidance but also does multi-robot mapping with limited resources in order not to compromise any agent during a mission.

\section{Methodology}
\label{methodology}

To optimize resource allocation in multi-agent systems, we use the RTD3 model for multi-agent goal-based exploration and collaborative mapping. We also implemented the GNN model, which does the goal allocation for each robot. This section details our implementation and highlights the key improvements and contributions.

\subsection{Refined TD3 Architecture}

\subsubsection{\textbf{Actor Network}}
The architecture of our Actor network is meticulously designed to cater to the complex state-action relationship. The Actor network embodies the policy function of the agent. Given a state $s$, the Actor predicts the optimal action $a$ that the agent should execute.

The input to the network consists of 4 neurons along with the lidar input. These four neurons make up the agent's state space - [$d_{goal}, \theta_{goal}, v, \omega $], which corresponds to the distance to the goal, angle to goal, linear velocity, and angular velocity respectively at each time step in the simulation. The first layer consists of 256 neurons, followed by the subsequent 128-neuron layer (compared to Cimurs et al. \cite{9645287} where they have 800 and 600 neurons, respectively). Both the layers thereafter have layer normalization, ReLU activation, and a dropout of 0.2. Lastly, the output layer produces 2 action values - $\left[v \left(\frac{(a + 1)}{2} \right), \omega \right]$. The Actor function $\pi$ can be represented as:
$$
a = \pi(s; \theta_\pi)
$$
where $\theta_\pi$ denotes the parameters of the Actor network and $s$ denotes the state space. Our model has one-fourth of the number of model parameters, which was essentially due to incorporating Reward Categorized Replay Buffer and Curriculum Learning. These concepts will be discussed in detail in further sections. 

\subsubsection{\textbf{Critic Network}}
The Critic's role is pivotal in value-based reinforcement learning. It estimates the Q-value of a given state-action pair, guiding the Actor's policy optimization. Our architecture employs a twin network approach Q1 \& Q2 to alleviate overestimation bias, a common pitfall in Q-learning methods. This is prevented by taking the $min(Q1, Q2)$.

The Critic network is similar to the Actor-network, except this has two separate pathways for both the state and action. The action, on the other hand, is transformed through a 256-neuron layer. The state and action representations are then concatenated and processed to produce the Q-value:
$$
Q(s, a; \theta_Q) = V_{a,s}
$$
where $Q(s, a; \theta_Q)$ represents the Q-value function parameterized by $\theta_Q$ and $V_{a,s}$ denotes the expected return when taking action $a$ in state $s$.

\subsection{Reward Categorized Replay Buffer}
The quality of stored experiences can profoundly influence an agent's learning trajectory. Our approach incorporates a Reward Categorized Replay Buffer (RCRB) that categorizes experiences based on their reward outcomes. This improvement in using a replay buffer makes training the Actor-Critic network faster than the original model. These categories include Positive Buffer ($B_+$), Neutral Buffer ($B_0$), and Negative Buffer ($B_-$). Depending on the reward value, the experience tuple — comprising of $(s, a, r, s')$ is sorted into one of these buffers.

\begin{figure}[]
  \centering
  \includegraphics[width=0.55\linewidth]{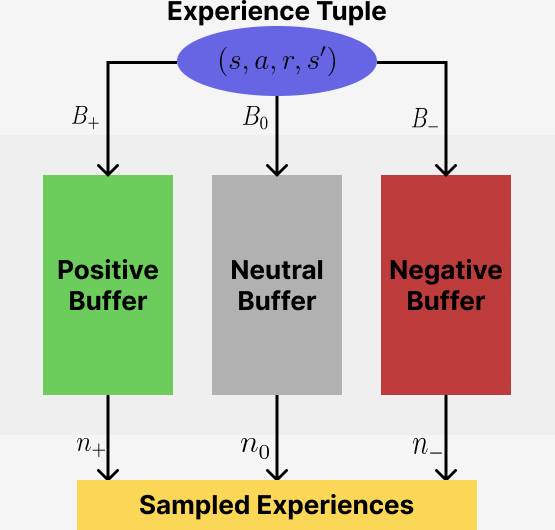}
  \caption{\textit{Representation of Reward Categorized Replay Buffer which accounts for Positive, Neutral, and Negative buffers based on the input experience replay}}
  \label{fig:rcrb}
\end{figure}

When training the agent, a balanced batch of experiences is required. The number of samples from each category is determined proportionally based on the current size of each buffer. This ensures that even rare occasions have a fair chance of being sampled, promoting a balanced learning experience for the agent as described below:
$$
n_{+} = \Bigl\lfloor \frac{len(B_{+})}{N} \times b \Bigl\rfloor ,
n_{0} = \Bigl\lfloor \frac{len(B_{0})}{N} \times b \Bigl\rfloor , 
n_{-} = b - n_{+} - n_{0}
$$
where $N$ is the total number of experiences in the buffer, and $b$ is the batch size for sampling. Figure \ref{fig:rcrb} describes the above-discussed approach.

This categorization offers several distinct advantages:
    \begin{enumerate}
        \item \textbf{Balanced Sampling:} Traditional buffers dilute sparse yet informative experiences in a vast sea of frequent outcomes. RCRB ensures a balanced representation, preventing bias towards commonly occurring rewards.
        \item \textbf{Efficient Learning:} By providing a representative sample of the environment's dynamics, RCRB can expedite the agent's learning, leading to faster convergence.
        \item \textbf{Optimal Memory Utilization:} RCRB's balanced storage prevents the over-representation of any specific reward category, ensuring efficient memory use.
    \end{enumerate}

\subsection{Training Pipeline - Refined TD3}

Central to the DREAM framework is the deployment of the RTD3 algorithm, a state-of-the-art approach for predicting continuous action spaces. The intricate design of our model, combined with our custom architectural choices, forms the backbone of our agent's learning mechanism.

\subsubsection{\textbf{Curriculum Learning}} 
\begin{figure}[h]
    \centering
    \begin{subfigure}{0.16\textwidth}
        \includegraphics[width=\textwidth]{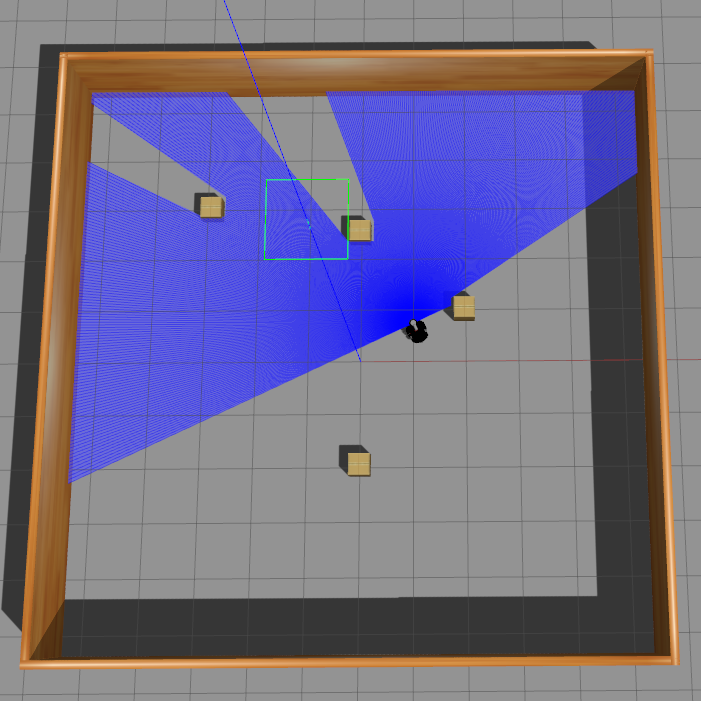}
    \end{subfigure}%
    \begin{subfigure}{0.16\textwidth}
        \includegraphics[width=\textwidth]{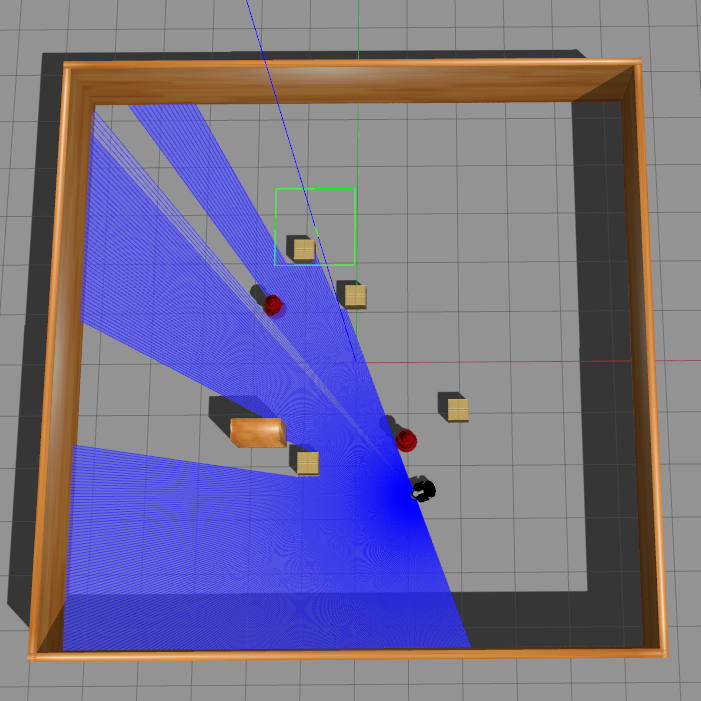}
    \end{subfigure}%
    \begin{subfigure}{0.16\textwidth}
        \includegraphics[width=\textwidth]{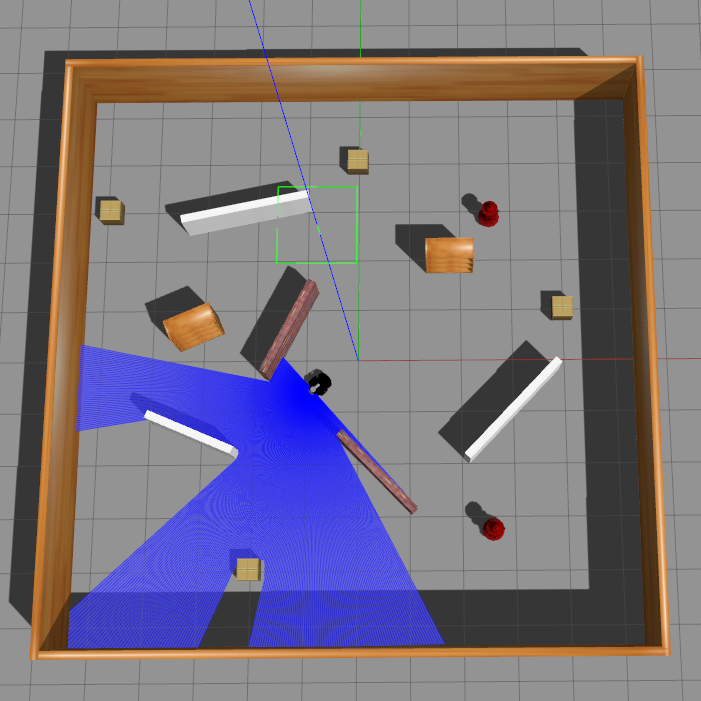}
    \end{subfigure}
    \caption{\textit{From left to right - The complexity of the environment and learning process increases as the training progresses. Initially, the obstacles are set at a fixed location. As time progresses, the number of obstacles grows, and their placement becomes random. A similar trend is observed for the robot's odometry, goal, and home positions.}}
    \label{fig:curriculum_learning}
\end{figure}

We commenced our training regimen with a basic environment setup involving one robot-goal pair with some obstacles at a fixed position. This enabled the model to grasp fundamental obstacle avoidance nuances. Eventually, the complexity of the environment was increased along with different obstacle poses during each episode to simulate real-world conditions. Figure \ref{fig:curriculum_learning} showcases this curriculum learning approach \cite{10.1145/1553374.1553380}.

Another aspect involved the goal and home positions spawned in close proximity to the robot's initiation point. As the model showcased increased competence, we expanded the distances to account for the challenge of prolonged journeys. This extension demanded the model to account for increased energy consumption, navigation challenges, and optimal path selection over elongated distances.

\subsubsection{\textbf{Critic Update}}
The target Q-values for the Critic's update are computed using a modified Bellman equation:
$$
Q_{target} = r + \gamma (1 - d) \min_{i=1,2} Q_{i,target}(s', \pi(s'))
$$
Here, $r$ represents the reward, $\gamma$ is the discount factor, $d$ is the termination flag and $s'$ is the subsequent state. The innovation lies in the use of the minimum Q-value from the two target Critic networks, a strategy that curbs overestimation bias. The primary Critic networks then predict the Q-values for the original actions. The Mean Squared Error between these predicted and target Q-values forms the loss, thereby guiding the Critic's optimization of parameters.

\begin{algorithm}[H]
\caption{Overview of the Improvised TD3 Approach}
\label{alg:modified_td3}
\begin{algorithmic}
\State Init: Actor, Critic networks; Target networks
\State Init: Reward Categorized Replay Buffer (RCRB)
\State Set: Curriculum Level $\leftarrow$ 1
\For{episode $\leftarrow$ 1 to N}
    \State Adjust environment by Curriculum Level
    \State Init environment; Obtain initial state $s$
    \While{!done}
        \State $a \leftarrow \pi(s; \theta_\pi) + \text{noise}$
        \State Execute $a$; Get $r$, $s'$, done
        \State Store ($s$, $a$, $r$, $s'$, done) in RCRB
        \State Sample from RCRB: states, actions, rewards, next\_states, dones
        
        \State $Q_{\text{target}} \leftarrow r + (1 - \text{done}) \times \gamma \times \min(Q_{\text{targets}})$
        \State Update Critic with MSE loss using $Q_{\text{target}}$
        
        \If{iteration mod delay $== 0$}
            \State Update Actor by maximizing Q-values
            \State Soft update target networks
        \EndIf
        \State $s \leftarrow s'$
    \EndWhile
    \State Modify battery parameters using actions, state
    \State Save networks at intervals or on improvement
    \If{Performance meets Curriculum Level}
        \State Increment Curriculum Level
    \EndIf
\EndFor
\end{algorithmic}
\end{algorithm}

\subsubsection{\textbf{Actor Update}}
Unlike the Critic's frequent updates, the Actor undergoes parameter refinement every two iterations, which is a design choice to ensure stability. The Actor's objective is to maximize the expected Q-values from one of the Critic networks. Given the Critic's representation of the environment's value structure, this maximization ensures the Actor's policy aligns with high-reward trajectories. The pseudocode for training the RTD3 approach is provided in \ref{alg:modified_td3}.

\subsubsection{\textbf{Reward Policy}}

The reward policy for this algorithm is designed according to the following function:
$$
r(s_t, a_t) = \begin{cases}
      r_{goal} & \textit{if } D_{goal} < D_{thresh}  \\
      r_{collision} & \textit{if } L_{min} < C_{thresh} \rightarrow \textit{collision} \\
      r_{nothing} & \textit{otherwise}
    \end{cases}
$$
Here, $r$ is the reward of the state-action pair $(s_t, a_t)$ at timestep $t$. If the distance to the goal $D_{goal}$ was less than the threshold $D_{thresh}$, $r_{goal} = 200$ was awarded. If the distance from the closest Lidar reading was less than the threshold $C_{thresh}$, it was considered as collision, and $r_{collision} = -100$ was awarded. In case the robot never encountered any of these conditions, $v - |\omega|$ was awarded to nudge the system to move more linearly than in the angular direction and ensure the agent learned to reach its goal in minimal steps.

\subsection{Obstacle Avoidance \& Navigation}

Since the robot used Lidar scan as input from the environment, it received high-dimensional sensory data reflecting the distances to nearby obstacles in a 180-degree view. This data can be utilized as the state space. However, since 180 is a high dimension for the model to learn, with most of it being redundant information, we represent the environment with only 30 values. This is achieved by dividing the 180-degree view into 30 segments.

The Lidar data provides the state space describing the immediate environment. Hence, the total state space for this algorithm consists of environment and robot states, totaling to 34, which becomes the input to the network. The agent trained in the RTD3 network takes action from a set of possible actions at every timestep to maximize cumulative reward.

The reward function provides positive feedback for actions that move the robot closer to its goal since it rewards more for moving linearly while avoiding obstacles. As the agent iterates through numerous episodes, it learns an optimal policy that maps states to actions, optimizing the trajectory to the goal while ensuring obstacle-free navigation. The network, when trained for a sufficient number of episodes, generalizes well to unseen environments, allowing the robot to navigate and avoid obstacles in novel scenarios using only its Lidar.

\subsection{Multi-Robot Decentralized Exploration \& Mapping}
We previously highlighted that our network was primarily designed to train a single agent to reach its target goal. This choice to focus on one agent rather than multiple agents simultaneously, was made to simplify the training process. This allowed the agent to deeply understand the environment's intricacies, including obstacles and the goal's location. Once the training was complete, we deployed this model on three robots, each navigating to their goals. With every episode, the environment, as well as the robot and goal poses, were randomized. The state space comprised 30 Lidar readings and the agent's state space. This setup enabled the robot to navigate around both static and dynamic obstacles (other robots).

\begin{figure}[]
  \centering
  \includegraphics[width=0.7\linewidth]{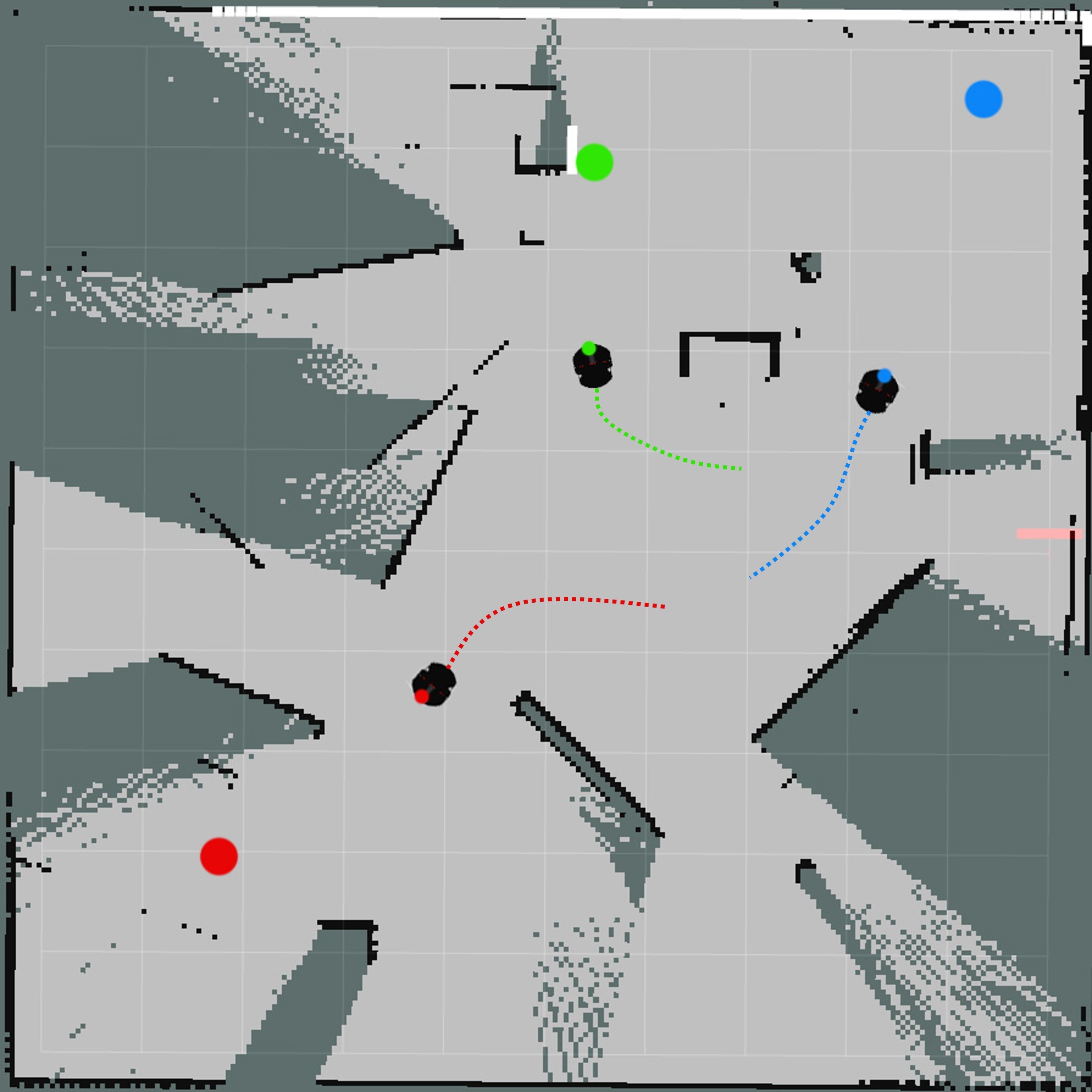}
  \caption{\textit{RViz2 showing three robots navigating to their respective goals along with mapping the environment collaboratively}}
  \label{fig:collab_map}
\end{figure}

We utilized \textit{slam toolbox} \cite{Macenski2021} ROS2 package \cite{doi:10.1126/scirobotics.abm6074} for mapping. These tools enabled collaborative mapping, using laser scans and odometry sensors to produce an occupancy grid map of the environment. Each robot was set to publish its parent TF transform as $/robot\_x/odom$. To facilitate collaborative mapping, we established additional TF layers, connecting these individual robots under a single parent. Specifically, $robot\_x/odom$ was made a child to $robot\_x/map$, and all such TFs were linked to the main $map$ frame. This structure allowed for map visualization in RViz2, as shown in Figure \ref{fig:collab_map}.

Initially, goals were randomly set before exploration began. Robots followed waypoints, moving towards the goals while mapping their surroundings. Once the goal was reached, that area was marked as explored, and a new goal was then set, thereby continuing this cycle for all robots.

\subsection{Battery Modeling}

In multi-agent systems, understanding and predicting battery behavior is critical. With the integration of energy management in the DREAM framework, simulating a battery's behavior allows for more informed decisions regarding energy allocation. In this work, we utilized the Doyle-Fuller-Newman (DFN) model to simulate and visualize battery discharge over time.

This DFN model provides a comprehensive representation of a lithium-ion battery by capturing the intricate electrochemical processes occurring within. Leveraging the PyBaMM library \cite{Sulzer2021}, we readily instantiate this model to serve as the foundation for our battery life simulation. Our target simulation duration was set to 900 seconds, equivalent to approximately 15 minutes of battery life for each robot.

The DFN model's default parameters are updated to reflect our calculated constant discharge current. This simulated battery was utilized in training the GNN model since it was one of the important factors for goal allocation. This module was not included in training the RL model as one of the rewards addressed this perspective of reaching the target quickly.

\subsection{GNN for Energy Management \& Goal Allocation}

In the realm of decentralized multi-robot systems, understanding and predicting the individual goals of robots are foremost. Inspired by the work of Li et al. \cite{li2021messageaware}, our approach leverages Graph Neural Networks (GNN) to model the relationship between different robots and predict their respective goals. This facilitates mission completion under constraint resource conditions while still being decentralized.

\subsubsection{Data Collection \& Preprocessing}

The framework is designed to simulate scenarios where three robots are assigned three distinct goals. The overarching objective is to pair each robot with a goal such that the cumulative energy consumption across all robots is minimized while maintaining the overall lifespan of the entire MRS.

\textbf{State space:} the state space include robot odometry $(r_x, r_y)$, home $(h_x, h_y)$, and goal $(g_x, g_y)$ locations.

\textbf{Energy Management Model:} The energy consumed by a robot is contingent on its linear or angular movement and the distance to the goal. Given $R$ robots and $G$ goals, we want to assign each robot to a goal. Each robot-goal pair has an associated energy consumption, which depends on the distance and orientation difference between the robot's current state and the goal. For each robot $i$ and goal $j$ pair, the energy $E_{ij}$ is computed based on the distance and orientation difference:
    $$
    E_{ij} = d_{target} \times battery_{straight} + \theta_{target} \times battery_{turn}
    $$
    where, $d_{target}$ is the normalized distance between the robot and the goal, $\theta_{target}$ accounts for the difference in orientation between the robot's current pose and its desired goal pose. $battery_{straight}$ = 10 and $battery_{turn} = 1.25 \times battery_{turn}$. The goal here is to find an assignment matrix $A$ of size $R \times G$ such that it minimizes the overall energy consumption
    $$
    A = \sum_{i=1}^{R} \sum_{j=1}^{G} E_{ij}
    $$
    where $E_{ij}$ represents the energy consumed by robot $i$ to reach goal. This methodology determines the permutation of robot-goal assignments that optimally reduces energy consumption at both the collective and individual levels. This is achieved by deploying a graph convolution neural network to predict the goal and home allocations, one after another.
    
\subsubsection{Network Architecture} \label{sec:network}

The architecture consists of multiple layers, beginning with a graph convolutional layer (GCN) followed by a fully connected layer. The network takes robot states, home locations, and goal positions as inputs. The inputs are concatenated and processed through two GCN layers. The first GCN layer takes the initial feature vector and transforms it into a higher-dimensional representation (128 dimensions) using the GCNConv operation, followed by a rectified linear unit (ReLU) activation function. Subsequently, the second GCN layer further refines the representation to a lower-dimensional space (64 dimensions) using the same convolutional operation and activation function. To prevent overfitting, dropout regularization is applied with a dropout rate of 0.5 during training. Finally, a fully connected layer maps the learned features to the prediction space, with the output reshaped to size $3 \times 2$, corresponding to the indices of goal and home locations for each robot. This architecture is designed to learn the mapping between the world states and the optimal goal allocation conditioned on robot battery levels.

\subsubsection{Training Pipeline - GCN} We use the Adam optimizer with a learning rate of 0.005 to update the network weights. The model parameters are initialized using the Xavier uniform initialization for the convolutional layers. This initialization method scales the weights based on the number of input and output units, which can aid in achieving a faster convergence. We design a special loss function to optimize energy consumption while maintaining the lifespan of each robot in the system. To this end, we compute the total energy of each robot on two journeys: from the current location to the goal and from the goal to home. Specifically, the network will optimize the total energy consumed by the whole system and also guarantee that each robot has enough battery to reach home. Any assignment that results in one or more stranded robots (low battery) during the mission, will be heavily penalized by our design. Consequently, the network will favor the optimal solutions for both global and individual levels.
 
\subsubsection{Model Insights}

The utilization of a graph-based paradigm offers notable advantages in terms of scalability and adaptability, particularly in the context of accommodating a variable number of robots within a system. In contrast to conventional methodologies which typically involve random goal assignment or demand computational complexity, the proposed model demonstrates the ability to efficiently predict individual goals for each robot based on their collective states.

\section{Experiments}
\label{experiments}

\subsection{System Setup}
We used a computer equipped with an NVIDIA GeForce RTX 3060 graphics card, with 32 GB of RAM, and 12th Gen Intel Core i7 12700K x 20 CPU running Ubuntu 22.04.3 and ROS2 Humble. We trained the Refined TD3 network using PyTorch \cite{paszke2019pytorch} in the Gazebo simulator \cite{1389727} for 7000 iterations.

\subsection{Training in Simulation}

The training pipeline was set up using ROS2, while Gazebo and RViz2 were used for visualization of laser scan, goal pose, home pose, robot odometry, and camera. The Pioneer P3-DX robot was used in the simulation. During the testing phase, three such robots were used with the same suite of sensors. 

When the training starts, Gazebo and RViz2 were disabled to accelerate the training process. One training episode is considered concluded when the robot reaches a goal, a collision is detected, or 250 steps are taken. As shown in figure \ref{fig:curriculum_learning}, the training was carried out in a simulated 10x10m sized environment as previously discussed.

We reward the robot for reaching the goal, penalize it for collisions, and give a minor reward otherwise. This encourages the robot to move, initially favoring forward motion over rotation. Despite early collisions, moving yields higher rewards than being stationary. The robot soon learns that avoiding obstacles, even if it means turning, is better than crashing. Hence, there's no need for a battery-saving penalty.

The state space doesn't consist of a home location since that is just another goal for the robot. This was confirmed during the evaluation phase as it is independent of the type of the goal. The model also doesn't account for dynamic obstacles, only a randomized environment on every reset. This has made the model robust enough to avoid dynamic obstacles like other robots. This was the primary reason not to train the three robots simultaneously. This was also confirmed in testing as the robots would swerve on encountering them in their path.

\subsection{Results}

To evaluate the effectiveness of our GNN-based goal allocation strategy versus a random allocation method, we introduce the Goal Allocation Evaluation Metric (GAEM). This metric is designed to measure the efficiency of different methods by comparing the overall energy consumption of robot teams in various environments.

\textbf{Metric Definition:} Given a multi-robot system (MRS), let \( E_{GNN} \) and \( E_{random} \) denote the energy consumption of robot \( r \) to reach goal \( g \) when goals are assigned using the GNN and random methods, respectively. The efficiency metric \( \epsilon \) for a particular environment is defined as:

\[
\epsilon = \frac{\sum_{r=1}^{R} E^{random}_{rg} - \sum_{r=1}^{R} E^{GNN}_{rg}}{\sum_{r=1}^{R} E^{random}_{rg}}
\]

A higher \( \epsilon \) indicates a more significant energy-saving using the GNN method compared to the random allocation. \( \epsilon \) offers a direct understanding of the percentage reduction in energy consumption using GNN over random goal allocation.

\begin{table}[]
    \centering
    \small 
    \begin{tabular}{|c|c|c|c|}
        \hline
        \textbf{Env} & \textbf{Avg \( E^{random} \)} & \textbf{Avg \( E^{GNN} \)} & \textbf{\( \epsilon \) (\% Reduction)} \\
        \hline
        1 & 124.67 & 93.4 & 25.08 \\
        2 & 136.01 & 102.56 & 24.59 \\
        3 & 157.32 & 120.98 & 23.09 \\
        4 & 163.71 & 123.16 & 24.76 \\
        5 & 122.83 & 96.67 & 21.29 \\
        \hline
    \end{tabular}
    \caption{\textit{Comparison of energy consumption for five different environments run five times each using random and GNN goal allocation methods.}}
    \label{tab:energy_comparison}
\end{table}

From Table \ref{tab:energy_comparison}, it is evident that this can be extended for various environments and a number of agents to showcase the strength and consistency of the GNN approach. The reduction in energy consumption not only improves the sustainability of the robot team but also enhances the efficiency and reliability of the operations on an individual level.

\section{Conclusion and Future Works}
\label{conclusion}

In this paper, we introduced \textit{DREAM} - a comprehensive framework that optimizes the allocation of vital resources in multi-agent systems. We successfully demonstrated the Refined TD3 model for obstacle avoidance, goal-based exploration and collaborative mapping for a multi-agent setup. This network had one-fourth number of model parameters due to the reward-categorized replay buffer and curriculum learning which created a significant impact on the rapid convergence of the model. We also presented a unique GNN model for goal allocation with resource constraints, since it is more optimized and scalable than a conventional approach.

Even though our model was trained in simulation, we believe this learned policy should be easily transferrable with minimal parameter changes on any type of robot with a Lidar sensor. Further, this model can be fine-tuned using pretraining to account for more velocity inputs for certain robots.

In this work, we have not addressed certain scenarios where robot(s) become incapable during a mission due to hardware or software issues. To alleviate this, one or more of the remaining agents would take over the task of the robot(s) which got compromised in order not to sacrifice the entire mission. This will be addressed in future work.

\bibliographystyle{IEEEtran}
\bibliography{refs}

\begin{thebibliography}{10}
\providecommand{\url}[1]{#1}
\csname url@rmstyle\endcsname
\providecommand{\newblock}{\relax}
\providecommand{\bibinfo}[2]{#2}
\providecommand\BIBentrySTDinterwordspacing{\spaceskip=0pt\relax}
\providecommand\BIBentryALTinterwordstretchfactor{4}
\providecommand\BIBentryALTinterwordspacing{\spaceskip=\fontdimen2\font plus
\BIBentryALTinterwordstretchfactor\fontdimen3\font minus
  \fontdimen4\font\relax}
\providecommand\BIBforeignlanguage[2]{{%
\expandafter\ifx\csname l@#1\endcsname\relax
\typeout{** WARNING: IEEEtran.bst: No hyphenation pattern has been}%
\typeout{** loaded for the language `#1'. Using the pattern for}%
\typeout{** the default language instead.}%
\else
\language=\csname l@#1\endcsname
\fi
#2}}

\bibitem{7986575}
C.~Mouradian, J.~Sahoo, R.~H. Glitho, M.~J. Morrow, and P.~A. Polakos, ``A
  coalition formation algorithm for multi-robot task allocation in large-scale
  natural disasters,'' in \emph{2017 13th International Wireless Communications
  and Mobile Computing Conference (IWCMC)}, 2017, pp. 1909--1914.

\bibitem{article}
D.~Drew, ``Multi-agent systems for search and rescue applications,''
  \emph{Current Robotics Reports}, vol.~2, 03 2021.

\bibitem{MultiRL}
L.~Busoniu, R.~Babuska, and B.~De~Schutter, ``A comprehensive survey of
  multiagent reinforcement learning,'' \emph{IEEE Transactions on Systems, Man,
  and Cybernetics, Part C (Applications and Reviews)}, vol.~38, no.~2, pp.
  156--172, 2008.

\bibitem{8352646}
A.~Dorri, S.~S. Kanhere, and R.~Jurdak, ``Multi-agent systems: A survey,''
  \emph{IEEE Access}, vol.~6, pp. 28\,573--28\,593, 2018.

\bibitem{s22218099}
\BIBentryALTinterwordspacing
S.~S. Binyamin and S.~Ben~Slama, ``Multi-agent systems for resource allocation
  and scheduling in a smart grid,'' \emph{Sensors}, vol.~22, no.~21, 2022.
  [Online]. Available: \url{https://www.mdpi.com/1424-8220/22/21/8099}
\BIBentrySTDinterwordspacing

\bibitem{LoweWTHAM17}
\BIBentryALTinterwordspacing
R.~Lowe, Y.~Wu, A.~Tamar, J.~Harb, P.~Abbeel, and I.~Mordatch, ``Multi-agent
  actor-critic for mixed cooperative-competitive environments,'' \emph{CoRR},
  vol. abs/1706.02275, 2017. [Online]. Available:
  \url{http://arxiv.org/abs/1706.02275}
\BIBentrySTDinterwordspacing

\bibitem{Yang}
\BIBentryALTinterwordspacing
Y.~Yang, R.~Luo, M.~Li, M.~Zhou, W.~Zhang, and J.~Wang, ``Mean field
  multi-agent reinforcement learning,'' \emph{CoRR}, vol. abs/1802.05438, 2018.
  [Online]. Available: \url{http://arxiv.org/abs/1802.05438}
\BIBentrySTDinterwordspacing

\bibitem{RAKESH2022108020}
\BIBentryALTinterwordspacing
S.~K. Rakesh and M.~Shrivastava, ``Performance analysis of fault tolerance
  algorithm for pattern formation of swarm agents,'' \emph{Knowledge-Based
  Systems}, vol. 240, p. 108020, 2022. [Online]. Available:
  \url{https://www.sciencedirect.com/science/article/pii/S0950705121011199}
\BIBentrySTDinterwordspacing

\bibitem{8511420}
P.~Pierpaoli, D.~Sauter, and M.~Egerstedt, ``Fault tolerant control for
  networked mobile robots,'' in \emph{2018 IEEE Conference on Control
  Technology and Applications (CCTA)}, 2018, pp. 374--379.

\bibitem{7101619}
C.~Di~Franco and G.~Buttazzo, ``Energy-aware coverage path planning of uavs,''
  in \emph{2015 IEEE International Conference on Autonomous Robot Systems and
  Competitions}, 2015, pp. 111--117.

\bibitem{energies}
K.~Nonoyama, Z.~Liu, T.~Fujiwara, M.~M. Alam, and T.~Nishi, ``Energy-efficient
  robot configuration and motion planning using genetic algorithm and particle
  swarm optimization,'' \emph{Energies}, vol.~15, p. 2074, 03 2022.

\bibitem{su141610056}
\BIBentryALTinterwordspacing
M.-F.~R. Lee and A.~Nugroho, ``Intelligent energy management system for mobile
  robot,'' \emph{Sustainability}, vol.~14, no.~16, 2022. [Online]. Available:
  \url{https://www.mdpi.com/2071-1050/14/16/10056}
\BIBentrySTDinterwordspacing

\bibitem{foerster2016learning}
J.~N. Foerster, Y.~M. Assael, N.~de~Freitas, and S.~Whiteson, ``Learning to
  communicate with deep multi-agent reinforcement learning,'' 2016.

\bibitem{veličković2018graph}
P.~Veličković, G.~Cucurull, A.~Casanova, A.~Romero, P.~Liò, and Y.~Bengio,
  ``Graph attention networks,'' 2018.

\bibitem{liu2020when2com}
Y.-C. Liu, J.~Tian, N.~Glaser, and Z.~Kira, ``When2com: Multi-agent perception
  via communication graph grouping,'' 2020.

\bibitem{7784001}
L.~Tai and M.~Liu, ``A robot exploration strategy based on q-learning
  network,'' in \emph{2016 IEEE International Conference on Real-time Computing
  and Robotics (RCAR)}, 2016, pp. 57--62.

\bibitem{chaplot2020learning}
D.~S. Chaplot, D.~Gandhi, S.~Gupta, A.~Gupta, and R.~Salakhutdinov, ``Learning
  to explore using active neural slam,'' 2020.

\bibitem{ramakrishnan2020occupancy}
S.~K. Ramakrishnan, Z.~Al-Halah, and K.~Grauman, ``Occupancy anticipation for
  efficient exploration and navigation,'' 2020.

\bibitem{8793769}
R.~Shrestha, F.-P. Tian, W.~Feng, P.~Tan, and R.~Vaughan, ``Learned map
  prediction for enhanced mobile robot exploration,'' in \emph{2019
  International Conference on Robotics and Automation (ICRA)}, 2019, pp.
  1197--1204.

\bibitem{9645287}
R.~Cimurs, I.~H. Suh, and J.~H. Lee, ``Goal-driven autonomous exploration
  through deep reinforcement learning,'' \emph{IEEE Robotics and Automation
  Letters}, vol.~7, no.~2, pp. 730--737, 2022.

\bibitem{10.1145/1553374.1553380}
\BIBentryALTinterwordspacing
Y.~Bengio, J.~Louradour, R.~Collobert, and J.~Weston, ``Curriculum learning,''
  in \emph{Proceedings of the 26th Annual International Conference on Machine
  Learning}, ser. ICML '09.\hskip 1em plus 0.5em minus 0.4em\relax New York,
  NY, USA: Association for Computing Machinery, 2009, p. 41–48. [Online].
  Available: \url{https://doi.org/10.1145/1553374.1553380}
\BIBentrySTDinterwordspacing

\bibitem{Macenski2021}
\BIBentryALTinterwordspacing
S.~Macenski and I.~Jambrecic, ``Slam toolbox: Slam for the dynamic world,''
  \emph{Journal of Open Source Software}, vol.~6, no.~61, p. 2783, 2021.
  [Online]. Available: \url{https://doi.org/10.21105/joss.02783}
\BIBentrySTDinterwordspacing

\bibitem{doi:10.1126/scirobotics.abm6074}
\BIBentryALTinterwordspacing
S.~Macenski, T.~Foote, B.~Gerkey, C.~Lalancette, and W.~Woodall, ``Robot
  operating system 2: Design, architecture, and uses in the wild,''
  \emph{Science Robotics}, vol.~7, no.~66, p. eabm6074, 2022. [Online].
  Available: \url{https://www.science.org/doi/abs/10.1126/scirobotics.abm6074}
\BIBentrySTDinterwordspacing

\bibitem{Sulzer2021}
\BIBentryALTinterwordspacing
V.~Sulzer, S.~G. Marquis, R.~Timms, M.~Robinson, and S.~J. Chapman, ``{Python
  Battery Mathematical Modelling (PyBaMM)},'' p.~14, June 2021. [Online].
  Available: \url{https://github.com/pybamm-team/PyBaMM}
\BIBentrySTDinterwordspacing

\bibitem{li2021messageaware}
Q.~Li, W.~Lin, Z.~Liu, and A.~Prorok, ``Message-aware graph attention networks
  for large-scale multi-robot path planning,'' 2021.

\bibitem{paszke2019pytorch}
A.~Paszke, S.~Gross, F.~Massa, A.~Lerer, J.~Bradbury, G.~Chanan, T.~Killeen,
  Z.~Lin, N.~Gimelshein, L.~Antiga, A.~Desmaison, A.~Köpf, E.~Yang, Z.~DeVito,
  M.~Raison, A.~Tejani, S.~Chilamkurthy, B.~Steiner, L.~Fang, J.~Bai, and
  S.~Chintala, ``Pytorch: An imperative style, high-performance deep learning
  library,'' 2019.

\bibitem{1389727}
N.~Koenig and A.~Howard, ``Design and use paradigms for gazebo, an open-source
  multi-robot simulator,'' in \emph{2004 IEEE/RSJ International Conference on
  Intelligent Robots and Systems (IROS) (IEEE Cat. No.04CH37566)}, vol.~3,
  2004, pp. 2149--2154 vol.3.

\end{thebibliography}
\end{document}